# Arguing for Decisions:
# A Qualitative Model of Decision Making

**Blai Bonet**
Dpto. de Computación
Universidad Simón Bolívar
Aptdo. 89000, Caracas 1080-A
Venezuela

**Hector Geffner***
Dpto. de Computación
Universidad Simón Bolívar
Aptdo. 89000, Caracas 1080-A
Venezuela

## Abstract

We develop a qualitative model of decision making with two aims: to describe how people make simple decisions and to enable computer programs to do the same. Current approaches based on Planning or Decision Theory either ignore uncertainty and tradeoffs, or provide languages and algorithms that are too complex for this task. The proposed model provides a language based on rules, a semantics based on high probabilities and lexicographical preferences, and a transparent decision procedure where reasons for and against decisions interact. The model is no substitute for Decision Theory, yet for decisions that people find easy to explain it may provide an appealing alternative.

## 1 INTRODUCTION

In this paper we develop a qualitative model of decision making with two aims: to describe how people make simple, everyday decisions and to enable computer programs to do the same. Current approaches based on Planning [Weld, 1994] or Decision Theory [Raiffa, 1970] either ignore uncertainty and tradeoffs, or provide languages and algorithms that are too complex for this task. The model proposed provides a simple language based on rules, a semantics based on high probabilities and lexicographical preferences, and a transparent decision procedure where reasons for and against decisions interact.

The model is closely related to other qualitative abstractions of Decision Theory (e.g., [Pearl, 1993; Boutilier, 1994; Wilson, 1995]), yet it introduces as-sumptions that aim to account for the way decisions are made in simple settings. In the proposed model, like in the findings of [Shafir et al., 1993; Hogarth and Kunreuther, 1995], the *reasons* for decisions play a central role. The result is an efficient, 'anytime' decision procedure, which is easy to justify and explain.

The paper is organized as follows. First we introduce the representation language (Section 2), the decision procedure (Section 3) and the semantics (Section 4). Then we discuss the relation to Decision Theory (Section 5), sensitivity issues (Section 6), extensions (Section 7), and related work (Section 8).

## 2 LANGUAGE

Models in the proposed framework contain four parts (see Fig. 1).

1) a set of input propositions and observations defining the possible input situations,

2) a set of goals and goals priorities defining the output situations,

3) a set of actions and action rules defining how input situations are mapped to output situations, and

4) plausibility measures defining the plausibility of the input situations

For example, a situation in which one has to decide whether to study for an exam or go to the beach can be modeled as:

```
study ∧ get-it    ⇒  pass-exam
go-beach ∧ ¬rain  ⇒  enjoy-beach
   unlikely rain  ;  plausible get-it
```

Here **study** and **go-beach** are the possible actions, **rain** and **get-it** are the input propositions and **pass-exam** and **enjoy-beach** are the *positive* goals in that order of importance.

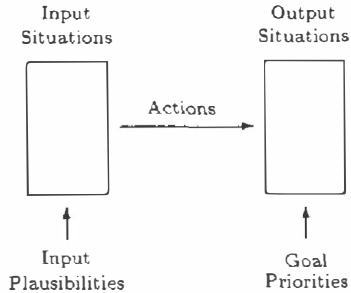

Figure 1: The Decision Situation

## 2.1 INPUT SITUATIONS

The *input situations* or *states* stand for the possible truth assignments to the *input propositions* and represent the *context* of the decision. With each input proposition $y$ we associate a *boolean variable* $Y$ such that $y$ stands for $Y = true$ and $\neg y$ stands for $Y = false$. We also use the notation Y to denote any of the literals $y$ or $\neg y$ associated with the variable $Y$, and $\sim$Y to denote its complement. The observations are input literals that have been found to be true.

## 2.2 GOALS

The *goals* stand for states of affairs that we care about. The *positive* goals are the ones that we want to achieve: getting a good job, enjoying a good day at the beach, watching a good movie, etc. The *negative* goals are the ones that we want to avoid: being dead, being thirsty, hurting people, missing an appointment, etc.

In this model, the goals are represented by literals (different from the input literals; yet see Section 7) denoted by symbols like x, x', .... The set of all goals is denoted by the letter $G$ while the set of positive and negative goals by $G^+$ and $G^-$ respectively. We use the words *goal literals* to refer to goals or their complements.

As the example above suggests, some goals are more important than others: getting a good job is more important than watching a good movie, being not dead is more important than missing an appointment, etc. We represent the *relative importance* of goals by integers: the higher the integer, the higher the importance of the goal. We call such integers the *priority* of the goals, and write $x \in G_i$ to say that the priority of goal x is $i$ (we will also write $x \in G_i^+$ or $x \in G_i^-$ when we want to make explicit the *polarity* of the goal as well). Schank and Abelson [1977] and Slade [1992] provide an interesting analysis of different types of goals (e.g., satisfaction, enjoyment, achievement, preservation, etc.) and their relative priorities.

Goal priorities are related to goal *utilities*, yet as degrees of *importance*, we assume that goal priorities combine as follows:

**Assumption 1** *Higher priority positive (negative) goals should be pursued (avoided) —even at expense of lower priority goals— except when success is deemed unlikely.*

This is a fundamental assumption in the model and says to focus on the actions that serve the most important goals, ignoring unlikely possibilities. This is not always a reasonable thing to do (see Section 6) but seems appropriate in the context of simple, everyday decisions. As we will see, this assumption will allow us cast the *decision process* as an *argumentation process* where reasons for and against decisions interact.

Because goals are important, we also assume that goals that are not said to be true *explicitly*,[1] are not true:[2]

**Convention 1** *Goals are assumed not true by default.*

Due to this convention, there is a difference in this model between declaring x as a positive goal and its complement $\sim$x as a negative goal. Even though in both cases we will try to achieve x and avoid $\sim$x, in the first case $\sim$x will be assumed true by default while in the second x will. In line with this convention we require that if x is a goal, $\sim$x is not.

## 2.3 ACTIONS AND ACTION RULES

The third component of the model are the *actions* and the *action rules*. Action rules map input situations (truth assignments to the input literals) to output situations (truth assignments to the goal literals). They are expressed by means of expressions of the form: $A \wedge C \Rightarrow x$, where $A$ is an action symbol, $C$ is a conjunction of input literals, and x is a goal literal (action symbols are distinct from input and output symbols).

Action rules are *default* rules in the sense that x is *normally* true after doing $A$ when $C$ is true. Each action rule has a *priority* or *strength measure* represented by a non-negative integer; the higher the number, the higher the priority. These priorities will be used to disambiguate conflicts among rules; e.g., to make a rule $A \wedge C \Rightarrow x$ override a conflicting rule $A \wedge C' \Rightarrow \sim x$ of lower priority. Unless otherwise specified, all rules are assumed to have priority zero (lowest priority).

Action rules which do not involve any actions, like knows-a-lot $\Rightarrow$ pass-exam, will be interpreted as abbreviations of rules involving the special action **do-nothing**, e.g., **do-nothing ∧ knows-a-lot ⇒ pass-exam**. The action **do-nothing** is assumed always present and represents the choice of not taking any (other) action. As for other actions, we can also have rules involving the symbol **do-nothing** explicitly (e.g., if a person is seriously injured and you do nothing, the person may die, etc.).

---

[1] Actually, there is no way to explicitly say that a goal is true in this language, yet see Section 7 for extensions that do.

[2] We distinguish 'assumptions' from 'conventions' to emphasize that the latter are just a matter of convenience; they are not built into the model like the former.

## 2.4 INPUT PLAUSIBILITIES

The last component of the model are the plausibilities of the input propositions. For that we allow the following type of statements for any input literal **Y**: '**likely Y**', '**plausible Y**' and '**unlikely Y**'. Intuitively, these statements rank the prior probability of **Y** in decreasing order, with the first and last denoting probabilities that are very close to 1 and 0 respectively. The meaning of these statements will be made precise in terms of Spohn's [1988] $\kappa$-functions. For the user's convenience we assume that:

**Convention 2** *When the input statements do not contain information about the plausibility of an input literal* **Y**, **Y** *is assumed* **plausible**.

## 2.5 EXAMPLE

The situation of going for the newspaper with or without the umbrella can be modeled in this language by means of action rules like:

$$
\begin{aligned}
\text{go-without-umbrella} &\Rightarrow \text{newspaper} \\
\text{go-with-umbrella} &\Rightarrow \text{newspaper} \\
\text{go-without-umbrella} \wedge \text{rain} &\Rightarrow \text{wet} \\
\text{go-with-umbrella} &\Rightarrow \text{carry}
\end{aligned}
$$

We also have to say that the possible actions are **go-with-umbrella** and **go-without-umbrella** and that the goals (and their polarities and priorities) are **carry** $\in G_1^-$, **wet** $\in G_2^-$ and **newspaper** $\in G_3^+$ (i.e., getting the newspaper is the most important goal, and avoiding getting wet is more important than avoiding having to carry an umbrella).

From the conventions above, it is implicit that **rain** is **plausible** (Convention 2), that each of the goals **newspaper**, **wet** and **carry** are true *only* when a rule asserting the goal is applicable (Convention 1), and that the action **do-nothing** does not achieve any goal.

## 3  REASONS FOR DECISIONS

We present now a *mechanism* for deciding which action to choose in a given context. The mechanism is based on the interplay of reasons. The procedure is efficient and easy to justify and explain. We start defining the reasons for decisions.

Basically, we will say that a positive (negative) goal **x** provides a *reason* for (against) action $A$ when the action $A$ contributes to the truth of **x**. The *polarity* of this reason is the polarity of the goal (positive or negative); the *importance* of the reason is the priority of the goal $(0, 1, \ldots, N)$; and the *strength* of the reason is the measure to which the action contributes to the truth of the goal.

More formally, let us say that a literal **Y** is *likely* when **Y** is an observation or when the information provided by the user contains **likely Y** or **unlikely ~Y**, and that **Y** is *unlikely* when its complement is likely, and *plausible* when **Y** is neither likely nor unlikely. Similarly, let us say that a *rule* $A \wedge C \Rightarrow \textbf{x}$ is *likely* when each conjunct in $C$ is likely, that is *unlikely* when some conjunct in $C$ is unlikely, and that is plausible when it is neither likely nor unlikely. Then, the *reasons* for decisions and their *strengths* are defined as follows:

**Definition 1** *A positive (negative) goal* **x** *provides a* strong *reason for (against) an action* $A$ *when some rule* $A \wedge C \Rightarrow \textbf{x}$ *is likely and no rule of the form* $A \wedge C' \Rightarrow \sim \textbf{x}$ *with equal or higher priority is either likely or plausible.*

**Definition 2** *A positive (negative) goal* **x** *provides a* weak *reason for (against) an action* $A$ *when it does not provide a likely reason for* $A$ *and yet some rule* $A \wedge C \Rightarrow \textbf{x}$ *is either likely or plausible, and no higher priority rule* $A \wedge C' \Rightarrow \sim \textbf{x}$ *is likely.*

**Definition 3** *A positive (negative) goal* **x** *provides a* empty *reason for (against) an action* $A$ *when it does not provide a strong or weak reason for (against)* $A$.

As an illustration, the goal **newspaper** provides a strong reason for **go-with-umbrella** and for **go-without-umbrella**; **wet** provides a weak reason against **go-without-umbrella**, and **carry** provides a weak reason against **go-with-umbrella**. Likewise, each of these goals provide empty reasons for or against **do-nothing**.

Clearly, decisions over a single goal can be taken by considering the strength and polarity of the reasons involved.

**Definition 4** *An action* $A$ *is* better than *an action* $B$ *over a* positive *goal* **x** *when* **x** *provides a stronger reason for* $A$ *than for* $B$.[3] *Likewise,* $A$ *is better than* $B$ *over a* negative *goal* **x** *if* **x** *provides a stronger reason against* $B$ *than against* $A$.

---
[3] Strong reasons are stronger than weak reasons, and weak reasons are stronger than empty reasons.

When there are many goals involved, the more important goals are considered first:

**Definition 5** *An action A is* better than *an action B, written $A > B$, when A is better than B over a goal x and B is no better than A over any goal x′ as important as or more important than x.*

The *overall best actions* are the actions that are no worse than any other action. We can test whether an action $A$ is better than an action $B$ by invoking the procedure better?$(A, B, i)$ that iteratively checks whether $A$ gets more compelling reasons than $B$ from goals in $G_i$, where $i$ is a priority level initially set to the top priority $N$. Indeed, better? must return **no** when some positive (negative) goal provides a stronger reason for $B$ ($A$) than for $A$ ($B$); **yes** when the opposite is true, and must call itself with the value of $i$ decreased when neither condition holds, returning **no** when $i < 0$. In the worst case, the complexity of this method is:

**Proposition 1** *The best actions can be computed in this way in time proportional to $\mathcal{A}^2 \times \mathcal{R}$, where $\mathcal{A}$ is the number of actions and $\mathcal{R}$ is the total number of rules.*

This complexity of this method is moderate, yet a more efficient procedure can be used when goals are *linearly ordered* (i.e., when no pair of goals have the same priority). If $\mathcal{A}$ stands for the set of all actions and $i$ is a priority level (initially set to $N$), select$(\mathcal{A}, i)$ can compute the best actions by retaining in $\mathcal{A}$, in each iteration, only the actions that get the strongest (weakest) reason from the single positive (negative) goal in $G_i$. This iteration terminates when $i < 0$ or when $\mathcal{A}$ becomes a singleton.

In the example above, do-nothing is pruned from $\mathcal{A}$ in the first iteration because it only gets an empty reason from the positive goal newspaper. In the second iteration, go-without-umbrella is also pruned as it gets a strong negative reason from the goal wet. The action go-with-umbrella then is the single best action as it is the only action left in $\mathcal{A}$.

### 3.1 EXAMPLE

Consider whether to approach some animal, e.g., a dog, that we don't know whether it is aggressive or not:

approach ⇒ satisfy-curiosity
approach ∧ aggressive ⇒ get-hurt

In this case, satisfy-curiosity is a low priority positive goal and get-hurt is a high priority negative goal. Given no other information, aggressive is assumed plausible by convention, and hence, the action approach gets a strong positive reason from satisfy-curiosity and a weak negative reason from get-hurt. Yet since get-hurt is the most important goal, the action approach is rejected for do-nothing which gets no (weak or strong) negative reasons. Note, however, that if observations lead to us revise the chances of aggressive to unlikely, the preferences would get reversed.

## 4 SEMANTICS

The semantics will make precise the meaning of all the constructs in the model and will provide an independent criterion for assessing the decision procedures above. In Decision Theory, actions $A$ are ranked by their expected utility:

$$EU(A) = \sum_s P(s) \sum_{s'} P_A(s'|s) \, U(s') \qquad (1)$$

where the $s$ and $s'$ denote the input and output states respectively. Here we use an approximation of this criterion with Spohn's [1988] $\kappa$-functions in place of probabilities, and lexicographical orderings in place of utility functions.

### 4.1 BELIEFS

Spohn [1988] describes a model for uncertain reasoning that combines the main intuitions underlying probability theory (context dependence, conditionalization, etc.) with the notion of plain beliefs (see also [Goldszmidt and Pearl, 1992]). Beliefs in Spohn's model are represented by means of a function $\kappa$ that assigns a non-negative integer measure to each world $w$ and that satisfies the following calculus:[4]

$$\kappa(p) \in [0, \infty] \, , \; \kappa(p) \; = \; \min_{w \models p} \kappa(w) \qquad (2)$$
$$\kappa(p \vee \neg p) = 0 \, , \; \kappa(p|q) = \kappa(p \wedge q) - \kappa(q) \qquad (3)$$

This calculus is structurally similar to the calculus of probabilities with products replaced by sums and sums replaced by minimizations. Spohn indeed showed that the first can be understood as an abstraction of the second with $\kappa$ measures standing for order-of-magnitude probabilities.

Lower $\kappa$ measures stand for higher probabilities and higher $\kappa$ measures stand for lower probabilities. Spohn indeed refers to the $\kappa$ measures as degrees of surprise or disbelief, hence regarding a proposition $p$ as *plausible* or *believable* when $\kappa(p) = 0$ and as *unlikely* or *disbelieved* when $\kappa(p) > 0$. In particular, since the axioms rule out two complementary propositions from being disbelieved at the same time, a proposition $p$ is *accepted* or *believed* when its negation is disbelieved (i.e., when $\kappa(\neg p) > 0$).

---
[4]$\kappa(p) = \infty$ when $p$ is unsatisfiable.

### 4.1.1   INPUT BELIEFS

We use Spohn's $\kappa$ functions to formalize the beliefs $\kappa(\mathbf{Y})$ in the inputs propositions and the beliefs $\kappa_A(\mathbf{x}|s)$ in the output propositions given an input state and action. For the inputs beliefs, we assume that:[5]

**Assumption 2** *Input variables are assumed to be independent.*

This means that, in analogy to probabilities, the belief in an input state is the aggregation of the beliefs in the input literals true in that state:

$$\kappa(\mathbf{Y}_1, \ldots, \mathbf{Y}_n) = \sum_{i=1}^{n} \kappa(\mathbf{Y}_i) \quad (4)$$

These beliefs in turn are provided by the user or assumed by default (Convention 2):

$$\kappa(\mathbf{Y}_i) = \begin{cases} 1 & \text{if unlikely } \mathbf{Y}_i \text{ or likely } \sim\mathbf{Y}_i \\ 0 & \text{otherwise} \end{cases} \quad (5)$$

These two equations determine the prior plausibility $\kappa(s)$ of any input state $s$ completely. The posterior plausibility $\kappa(s|\mathbf{obs})$ given a set of observations (input literals) can be derived from (3) as $\kappa(s|\mathbf{obs}) = \kappa(s) - \kappa(\mathbf{obs})$ if $s$ satisfies $\mathbf{obs}$, and $\infty$ otherwise.

### 4.1.2   GOAL BELIEFS

For any goal literal $\mathbf{x}$, the plausibility of $\mathbf{x}$ given an action $A$ and an input state $s$ is expressed by the equation:

$$\kappa_A(\mathbf{x}) = \min_s (\kappa(s) + \kappa_A(\mathbf{x}|s)) \quad (6)$$

which is the qualitative version of the equation $P_A(\mathbf{x}) = \sum_s P(s) P_A(\mathbf{x}|s)$.

From Equations 4 and 5, we know how to determine the plausibilities $\kappa(s)$; we are thus left to determine the conditional plausibilities $\kappa_A(\mathbf{x}|s)$. These plausibilities will be extracted from the rules that are applicable in the state $s$ that relate $A$ to $\mathbf{x}$.

Let $r(A)$ denote the set of action rules involving the action $A$. Then we say that the input state $s$ *supports* a literal $\mathbf{x}$ when there is a rule in $r(A)$ whose consequent is $\mathbf{x}$, whose conditions are true in $s$, and for which all conflicting rules in $r(A)$ whose conditions are true in $s$ have equal or lower priority.

The plausibilities $\kappa_A(\mathbf{x}|s)$ are then defined to capture Convention 1 (goals are assumed false by default) and

---

[5] We will show how to relax this assumption in Section 7.

---

the intuition that supported literals should be either likely or plausible (they have a justification):

$$\kappa_A(\mathbf{x}|s) = \begin{cases} 0 & \text{when } \mathbf{x} \text{ is supported by } s, \text{ or when} \\ & \sim\mathbf{x} \text{ is a goal not supported by } s \\ 1 & \text{otherwise} \end{cases}$$
(7)

Equations 4–7 determine the measures $\kappa_A(\mathbf{x})$ for any value $\mathbf{x}$ and any action $A$. When input observations $\mathbf{obs}$ are gathered, the conditional measure $\kappa_A(\mathbf{x}|\mathbf{obs})$ can be obtained by replacing the prior plausibility $\kappa(s)$ in Equation 6 by the posterior plausibility $\kappa(s|\mathbf{obs})$.

### 4.1.3   EXAMPLE

We illustrate these definitions in the newspaper example. Because the only input variable is rain, which is assumed plausible by default, the input states are $s = \{\text{rain}\}$ and $s' = \{\neg\text{rain}\}$ with $\kappa(s) = \kappa(s') = 0$.

If $A$, $B$ and $C$ denote the actions go-without-umbrella, go-with-umbrella and do-nothing, the literals supported by each action in each input situation are:

|   | $s = \text{rain}$ | $s' = \neg\text{rain}$ |
|---|---|---|
| $A$ | $\{\text{news},\text{wet},\neg\text{carry}\}$ | $\{\text{news},\neg\text{wet},\neg\text{carry}\}$ |
| $B$ | $\{\text{news},\neg\text{wet},\text{carry}\}$ | $\{\text{news},\neg\text{wet},\text{carry}\}$ |
| $C$ | $\{\neg\text{news},\neg\text{wet},\neg\text{carry}\}$ | $\{\neg\text{news},\neg\text{wet},\neg\text{carry}\}$ |

From this table and Equations 4–7, the plausibilities of all goal literals can be computed; e.g., $\kappa_A(\text{wet}) = \min\{\kappa(s) + \kappa_A(\text{wet}|s), \kappa(s') + \kappa_A(\text{wet}|s')\} = \min\{0 + 0, 0 + 2\} = 0$.

## 4.2   PREFERENCES OVER ACTIONS

To rank the actions, we define the *qualitative utility* of a goal $\mathbf{x}$, written $u(\mathbf{x})$, as:

$$u(\mathbf{x}) = polarity(\mathbf{x}) \times priority(\mathbf{x}) \quad (8)$$

Namely, for $\text{wet} \in G_2^-$, $u(\text{wet}) = -2$, while for $\text{newspaper} \in G_3^+$, $u(\text{newspaper}) = 3$.

Provided with these measures, we could define the qualitative *expected utility of actions* relative to a goal $\mathbf{x}$, following the methods in [Pearl, 1993] or [Wilson, 1995], e.g., setting it to $\max(0, u(\mathbf{x}) - \kappa_A(\mathbf{x}|\mathbf{obs}))$ when $\mathbf{x}$ is positive. The problem with these schemes is that they impose a very strong requirement on the way utility measures are encoded so they can be added up, in the same scale, with $\kappa$ measures (see [Wilson, 1995]).

Here we take a different approach which does not require goal priorities and plausibility judgements to be so calibrated. In the proposed scheme, only two things matters: the *ordinal* ranking of goals, and whether goals are deemed likely, unlikely or plausible. This is done by defining the *qualitative belief* in a goal literal x as:

$$b_A(\mathbf{x}) = \kappa_A(\neg \mathbf{x}|\text{obs}) - \kappa_A(\mathbf{x}|\text{obs}) \qquad (9)$$

and defining the *qualitative rank* of an action $A$ relative to a goal x as:[6]

$$Q_{\mathbf{x}}(A) = sign(u(\mathbf{x})) \times sign(b_A(\mathbf{x})) \qquad (10)$$

In other words, an action has a *positive rank* relative to a goal x ($Q_{\mathbf{x}}(A) = 1$) when it's likely to make the positive (negative) goal x true (false); it has a *negative rank* ($Q_{\mathbf{x}}(A) = -1$) when it's likely to make the positive (negative) goal x false (true); and it has an *null* rank otherwise ($Q_{\mathbf{x}}(A) = 0$). Clearly,

**Definition 6** *An action $A$ is preferred to an action $B$ over a goal x if $Q_X(A) > Q_X(B)$.*

In the presence of multiple goals, this ordering is extended by considering more important goals first:

**Definition 7** *An action $A$ is preferred to an action $B$, written $A \succ B$, if $A$ is preferred to $B$ over some goal x, and $B$ is not preferred to $A$ over any goal x' with equal to or higher priority than x.*

The overall optimal actions determined by this preference relation are closely related to the actions that result from the decision procedures based on rules (Section 3). Indeed, if we say that a theory is *positive* when the rules do not involve negative literals in their bodies we get that:

**Proposition 2** *The decision procedures based on interaction of reasons are sound and complete for positive theories.*

The condition of positivity is required because the procedures do not reason by cases and thus cannot properly handle pairs of rules like $A \wedge p \Rightarrow \mathbf{x}$ and $A \wedge \neg p \Rightarrow \mathbf{x}$. Under this condition, Proposition 1 guarantees that the problem of identifying the best actions can be computed efficiently.

### 4.2.1 EXAMPLE

The table below summarizes the qualitative rank of the actions relative to each of the goals **newspaper**, **wet** and **carry**:

| Action | ⟨ n, w, c ⟩ |
|---|---|
| go-without-umbrella: | ⟨+1, +0, +1⟩ |
| go-with-umbrella: | ⟨+1, +1, −1⟩ |
| do-nothing: | ⟨−1, +1, +1⟩ |

---

[6]The function *sign* maps positive numbers into 1, negative numbers into −1, and 0 into 0.

The preferences among the actions are easy to visualize as they correspond to the lexicographical preferences among their corresponding vectors (in this case, no pair of goals have the same priority). The table makes evident that `go-with-umbrella` is the best action in this case. On the other hand, if `rain` were unlikely, the first entry below `wet` would become +1, and the best action would become `go-without-umbrella`.

## 5 RELATION TO DECISION THEORY

The optimal action $A$ from a decision theoretic point of view is the action that maximizes the expected utility (Equation 1). The model can be understood as assuming that the utility function $U(s)$ is *additively decomposable* as $U(s) = \sum_{\mathbf{x} \in s} U(\mathbf{x})$, where x is the value of variable $X$ in the output situation $s$, and that the utility of positive (negative) goals x is a fixed positive (negative) value $U_{\mathbf{x}}$ and the utility for the negation of a goal is 0. From these assumptions, it is possible to show that Equation 1 can be expressed as:

$$EU(A) = \sum_{\mathbf{x} \in G} U_{\mathbf{x}} P_A(\mathbf{x}) \qquad (11)$$

Furthermore, the model assumes that terms $U_{\mathbf{x}} P_A(\mathbf{x})$ make terms $U_{\mathbf{x}'} P_A(\mathbf{x}')$ negligible when $P_A(\mathbf{x}) \gg P_A(\mathbf{x}')$ ('unlikely scenarios are ignored') and that $|U_{\mathbf{x}}| \gg |U_{\mathbf{x}'}|$ when the priority of x is higher than the priority of x' ('low priority goals are traded by higher priority goals').

## 6 SENSITIVITY ISSUES

It is not hard to think of cases where the assumptions embedded in this model are not reasonable. Consider for example a situation in which a patient has a very serious disease which if not treated will result in his death. Moreover, there is only one possible treatment and such treatment does not always work, and in all cases it has undesirable side-effects like loosing hair, vomiting, etc.

$$\begin{aligned}
\text{do-nothing} &\Rightarrow \text{death} \\
\text{treatment} \wedge \neg\text{effective} &\Rightarrow \text{death} \\
\text{treatment} &\Rightarrow \text{side-effects}
\end{aligned}$$

Here the goals **death** and **side-effects** are both negative and the first is significantly more important than the second.

The atom **effective** provides the condition under which the treatment works. If the prior plausibility measure of **effective** is 0 (the treatment can plausibly work) the model recommends treatment. Yet if the prior plausibility measure of effective is 1 (the treatment most likely will not work) the model will recommend to do nothing (i.e., the action do-nothing will be preferred to treatment).

One way to look at the second scenario is that the model prefers the lottery 'certain death' to the lottery 'certain side-effects and very likely death'. This preference, which is not reasonable, results from *regarding unlikely scenarios as impossible ones*. This assumption, in cases where important goals are at stake, is actually far from appropriate.

We can measure though how robust an optimal decision is by considering how it is affected by changes in the input parameters (goal priorities and input plausibilities).

Let us say that a goal x *justifies* the preference of action $A$ over action $B$ if $A$ is preferred to $B$ over x and yet $A$ and $B$ are equally preferred over all goals with *higher* priority than x.

For example, when the treatment is unlikely to work, the goal that justifies the decision do-nothing over treatment is side-effects. On the other hand, when the treatment can plausibly work, do-nothing becomes inferior to treatment because of the goal death. Since death is considerably more important than side-effects the proper selection of the parameter $\kappa(\text{effective})$ in this case is critical. More generally, when minimal changes in an input parameter lead to abrupt changes in the importance of the goals that are obtained the optimality of the decisions need to be reconsidered. This critical tradeoff can be detected in this model, yet the same model is not sufficiently expressive to resolve them. Often, however, there may be no reasonable ways for resolving such tradeoffs.

## 7 EXTENSIONS

The expressive power of the model is limited yet there are a number of extensions that can be accommodated.

First, we can relax the assumptions that input variables be independent by accommodating *input rules* in addition to action rules. These input rules will impose a causal structure on the input variables which can be interpreted as in [Goldszmidt and Pearl, 1992] or [Geffner, 1996a]. Semantically the only difference is in the determination of the plausibilities of the input state $\kappa(s)$.

Second, we can interpret the input and output situations as referring to the state of the world before and after the action. The values of variables that occur in both the inputs and the outputs can then be assumed to persist by default [Gelfond and Lifschitz, 1993; Geffner, 1996b]. This can enable us to express *sequential* decision problems, where the choice of optimal actions is replaced by the choice of optimal action *sequences*.

In many cases, we may also need a way for representing and aggregating preferences among equally important goals. That is, two goals may be equally important and yet one may be preferred to the other; e.g., going to see the 'Knicks' vs. going to see the 'Mets'. A possible approach in this case is to express these preferences by means of integers and to aggregate such preferences by some form of weighted addition according to whether the goals are rendered likely, plausible or unlikely by the actions.

## 8 RELATED WORK

The proposed model for decisions is related to other qualitative abstractions of decision theory and to informal models of decisions based on the interplay of reasons.

Qualitative models of decision making have received considerable attention in recent years [Pearl, 1993; Boutilier, 1994; Dubois and Prade, 1995; Wilson, 1995]. All of these proposals have in common the use of qualitative measures for representing preferences and beliefs, yet compared to this work, few have placed emphasis on *modeling* (yet, see [Brewka and Gordon, 1995]) and in the mechanisms for *computing* and *explaining* decisions.

The work differs from [Pearl, 1993] and [Wilson, 1995] in the way utility ranks and $\kappa$ measures are combined. Pearl and Wilson assume that these measures are calibrated so that they can be added up in the same scale. Thus, a likely world with utility rank 1 is deemed as good as an unlikely world with utility measure 2. Our choice here is different: our priority measures are completely *ordinal* and represent the importance of goals. Our criterion is that most important goals dominate less important goals except when the former are unlikely to be realized.

The two criteria can be usefully contrasted in the simple case in which there is a single positive goal x involved. This scenario can be expressed in Pearl's and Wilson's framework by partioning the set of worlds into two sets: the worlds $w^+$ that satisfy x, which get a utility rank $\mu(w^+) = 1$, and the worlds $w^-$ that do not satisfy x, which get a utility rank $\mu(w^-) = 0$. A weakness of Pearl's and Wilson's scheme is that they fail to prefer actions $A$ that make x *likely* ($\kappa_A(\neg \mathbf{x}) > 0$) to actions $B$ that make x just *plausible* ($\kappa_B(\mathbf{x}) = \kappa_B(\neg \mathbf{x}) = 0$). Both actions get actually the same expected utility rank in their scheme. Inter-estingly this is not solved when the worlds $w^-$ that do not satisfy the goal are assigned a negative utility rank $\mu(w^-) = -1$. In that case, Pearl's and Wilson's schemes will label the actions $B$ that make the goal x plausible, *ambiguous.* We, on the other hand, rank such actions below the actions $A$ that make x likely, and above the actions $C$ that make x unlikely.

The procedures considered in Section 3 are related also to informal models of decision based on the interplay of reasons. For example, when one action $A$ gets either positive or negative (non-empty) reasons such that no other action gets (non-empty) reasons of the same importance, the action $A$ can immediately be accepted, if the reasons are positive, and rejected, if the reasons are negative. These type of situations, where there are clear and compelling reasons for accepting or rejecting decisions, seem to be the ones people feel most comfortable with and have been studied in [Shafir *et al.*, 1993].

### Acknowledgments

We want to thank the anonymous UAI reviewers for useful comments.